\documentclass{article}

\usepackage{microtype}
\usepackage{graphicx}
\usepackage{subcaption}
\usepackage{booktabs} 

\usepackage{hyperref}

\usepackage[preprint]{icml2026}

\usepackage{amsmath}
\usepackage{amssymb}
\usepackage{mathtools}
\usepackage{amsthm}

\usepackage[capitalize,noabbrev]{cleveref}

\theoremstyle{plain}

\theoremstyle{definition}

\theoremstyle{remark}

\usepackage[disable,textsize=tiny]{todonotes}

\icmltitlerunning{ISAAC: Auditing Causal Reasoning in Deep Models for Drug–Target Interaction Prediction}

\begin{document}

\twocolumn[
  \icmltitle{ISAAC: Auditing Causal Reasoning in Deep Models \\
    for Drug–Target Interaction Prediction}



  \icmlsetsymbol{equal}{*}

  \begin{icmlauthorlist}
  \icmlauthor{Barbara Tarantino}{pv}
  \icmlauthor{Sun Kim}{snu}
  \icmlauthor{Yijingxiu Lu}{snu}
  \icmlauthor{Paolo Giudici}{pv}
  \end{icmlauthorlist}

  \icmlaffiliation{pv}{Department of Economics and Management, 
  University of Pavia, Pavia, Italy}
  \icmlaffiliation{snu}{Department of Computer Science and Engineering, 
  Seoul National University, Gwanak-gu 08826, Korea}

  \icmlcorrespondingauthor{Paolo Giudici}{paolo.giudici@unipv.it}
  \icmlkeywords{Machine Learning, ICML}

  \vskip 0.3in
]



\printAffiliationsAndNotice{}  

\begin{abstract}
Deep learning models often achieve high predictive performance on scientific benchmarks, yet it remains unclear whether such performance reflects mechanistically grounded reasoning or reliance on spurious correlations. This ambiguity is especially problematic in scientific domains such as drug--target interaction prediction, where valid outputs are expected to depend on underlying molecular mechanisms rather than on surface-level statistical patterns.
We introduce ISAAC, a post-hoc structural auditing framework based on controlled input-level interventions, enabling a principled comparison of model responses to mechanistic and spurious perturbations to assess reliance on structurally meaningful input components beyond predictive accuracy.
Applying \textsc{ISAAC} to drug--target interaction prediction, we show that models with comparable predictive performance exhibit substantial and reproducible differences in reasoning scores that remain undetected under accuracy-based evaluation, motivating the use of post-hoc structural auditing alongside standard performance metrics.
\end{abstract}

\section{Introduction}

Deep learning models have achieved strong predictive performance across a wide range of scientific tasks, including molecular modeling, physical reasoning, and biological prediction. In practice, high accuracy is often interpreted as evidence that a model has learned to reason about the underlying problem. This assumption is misleading. Predictive accuracy measures agreement with observed labels, but it does not reveal how predictions are produced or whether they depend on meaningful causal structure.

In many scientific settings, correct predictions can be obtained without reasoning about the underlying mechanisms. 
Models may exploit statistical regularities, dataset-specific artifacts, or shortcut features that correlate with the target in-distribution but are not mechanistically relevant. As a result, two models with similar accuracy can exhibit fundamentally different reasoning behavior, and models that perform well on standard benchmarks may fail under intervention or distribution shift \citep{ilyas2019adversarial, taori2020measuring}. In scientific applications, this gap is particularly problematic, as predictions that are accurate yet mechanistically invalid can lead to misleading interpretations or unsafe downstream decisions.

This distinction highlights a limitation of accuracy-based evaluation. Accuracy is an observational metric that assesses performance on a fixed data distribution. 
Reasoning, by contrast, can be probed through interventional analysis. Assessing whether a model relies on mechanistically meaningful structure requires evaluating how its predictions change when relevant components of the input are actively perturbed. Without such probes, accuracy alone cannot distinguish causal dependence from spurious association \citep{pearl2009causality}.

Existing approaches provide only partial insight into this problem. Interpretability methods offer correlational explanations of model predictions, but explanation stability does not imply causal validity \citep{adebayo2018sanity}. Robustness and invariance-based approaches focus primarily on training objectives and generalization guarantees, rather than on explicitly evaluating the reasoning behavior of trained models \citep{peters2016causal}. As a result, there is no standard evaluation framework that directly tests whether predictions are driven by structurally meaningful input components rather than by superficial correlations.

Motivated by this gap, we introduce ISAAC (Intervention-based Structural Auditing Approach for Causal Reasoning), a post-hoc structural auditing framework based on controlled input-level interventions, designed to evaluate prior-relative structural sensitivity independently of predictive accuracy.
ISAAC treats reasoning as a behavioral property of a trained model, conditional on its predictive regime, and assesses it by probing frozen networks with structured interventions, where structural sensitivity is operationalized as differential response to
structurally aligned versus misaligned input-level interventions.
By comparing model responses to mechanistic interventions and matched spurious interventions, ISAAC tests whether predictions are sensitive to structurally meaningful input components or dominated by non-mechanistic correlations.

We instantiate ISAAC in the context of Drug--Target Interaction (DTI) prediction, a setting in which valid predictions are expected to depend on localized protein--ligand interactions and where biologically grounded interventions can be defined. Across multiple sequence-based DTI architectures spanning a range of inductive biases, we find that models with comparable predictive performance differ substantially in their responses to structured perturbations. These differences reveal differences in reasoning behavior that remain undetected under accuracy-based evaluation.

Overall, this work demonstrates that predictive accuracy alone is insufficient to characterize model reasoning in scientific machine learning. ISAAC introduces a principled, intervention-based auditing framework that
makes it possible to assess prior-relative structural sensitivity of trained
models beyond observational performance.

\section{Background and Related Work}

\subsection{Accuracy-Based Evaluation in Scientific Machine Learning}

In many scientific applications, model performance is still primarily assessed through predictive accuracy on held-out data. While useful, this criterion alone provides limited information about how predictions are obtained. A substantial body of work has shown that deep models can reach high accuracy by exploiting correlations that are predictive in-distribution but unrelated to the underlying mechanisms of interest for the task. Such shortcut features often support strong benchmark performance while remaining fragile under perturbations or distribution shift \citep{ilyas2019adversarial, geirhos2020shortcut}. These behaviors have been observed across modalities and tasks, and can persist even when models are optimized for robustness or calibrated performance \citep{taori2020measuring, kirichenko2020break}.

Several approaches have attempted to reduce shortcut reliance by promoting invariance across environments or perturbations. Invariant Risk Minimization formalizes the principle that predictors should rely on features whose relationship with the target remains stable across multiple environments \citep{arjovsky2019invariant}. Earlier work on invariant prediction relates such stability to causal structure \citep{peters2016causal}. While these frameworks provide valuable theoretical insights into causal generalization, they are primarily concerned with training objectives and guarantees on learned predictors. They do not directly address how a trained model behaves at inference time, nor whether its predictions depend on invariant or mechanistically meaningful features.

Related limitations arise in the context of interpretability and explainability. Feature attribution methods, including saliency maps, gradient-based techniques, local surrogate models, and Shapley-based explanations, aim to provide insight into individual predictions \citep{simonyan2014deep, sundararajan2017axiomatic, ribeiro2016lime, lundberg2017unified}. Although widely used for transparency and debugging \citep{doshi2017accountability}, these methods remain fundamentally correlational. Empirical studies have shown that attribution maps can remain largely unchanged even when model parameters are randomized or when predictions are driven by non-causal features \citep{adebayo2018sanity}. As a result, neither predictive accuracy nor explanation-based analyses provide a direct protocol for explicitly assessing whether trained models rely on meaningful input structure when producing predictions.

\subsection{Intervention-Based Approaches to Reasoning Evaluation}

Purely observational evaluation is often insufficient to distinguish meaningful reasoning behavior from spurious associations. Causal inference addresses this limitation by providing a formal framework for distinguishing associative from causal relationships through the analysis of interventions rather than observational regularities alone \citep{pearl2009causality}. This perspective has motivated extensive work on causal discovery, causal prediction, and causal representation learning. Invariant prediction methods, for example, exploit stability across interventions to identify causal relationships \citep{peters2016causal}, while structural causal models provide a formal language for representing and reasoning about underlying mechanisms \citep{pearl2009causality}.

More recent work on causal and disentangled representation learning emphasizes learning representations aligned with generative factors and causal mechanisms \citep{locatello2020weakly, scholkopf2021toward}. While principled, these approaches primarily focus on training-time objectives and representation properties, rather than on auditing the reasoning behavior of complex, pretrained models operating as black boxes.

A related line of work studies post-hoc causal analysis of trained neural networks through interventions on internal model components, including mediation-based analyses and causal abstraction frameworks based on interchange interventions \citep{Geiger2021causalabstractions, vig2020causal}. These approaches provide insight into internal causal mechanisms, but typically require access to model internals and assume a meaningful decomposition of latent variables or internal representations.

Intervention-based reasoning evaluation has also emerged in other domains, notably in large language models, where controlled perturbations and benchmark construction reveal gaps between apparent accuracy and underlying reasoning behavior \citep{wei2023reimagine, zhang2025llmscan}. Despite these advances, existing intervention-based approaches do not provide a principled post-hoc protocol for reasoning evaluation under a controlled predictive regime. In particular, they fail to systematically contrast sensitivity to structurally aligned versus misaligned input perturbations while jointly controlling for intervention magnitude, operator, and predictive performance, limiting their ability to audit whether pretrained scientific models preferentially rely on mechanistically meaningful input structure rather than spurious correlations.

\section{Methodology}

ISAAC provides a protocol for evaluating whether the predictions of a trained deep model
are supported by structurally meaningful reasoning rather than superficial statistical associations.
Reasoning is operationalized as a behavioral property of a fixed predictive system,
revealed through structured input-level interventions.


\subsection{Reasoning as Interventional Behavior}

Let $f_\theta \colon \mathcal{X} \rightarrow \mathcal{Y}$ denote a predictive model with fixed parameters $\theta$, selected prior to auditing. The input space $\mathcal{X}$ may correspond to sequences, graphs, images, or structured records, while the output space $\mathcal{Y}$ denotes a real-valued prediction quantity. Throughout auditing, the model is treated as a fixed predictive system.

We assume access to a finite auditing set $\mathcal{D} = \{x_i\}_{i=1}^N \subset \mathcal{X}$. Importantly, $\mathcal{D}$ is disjoint from the data used for training and validation, but does not necessarily coincide with the standard test set; instead, it refers to the subset of available data on which interventional analysis can be meaningfully performed.

Reasoning is then probed through externally specified input-level interventions. An intervention is defined as a deterministic operator $\mathcal{I} \colon \mathcal{X} \rightarrow \mathcal{X}$ that modifies selected components of an input while leaving the internal computation of the model unchanged. For an input $x \in \mathcal{D}$, the intervened input is denoted by $x^{\mathcal{I}} = \mathcal{I}(x)$. Interventions are introduced to induce controlled structural variation, independently of the data-generating distribution. The auditing set $\mathcal{D}$ is restricted to inputs admitting a structured representation that supports well-defined interventions; inputs lacking such structure are excluded.

Under this setup, ISAAC characterizes model behavior conditional on the learned parameters $\theta$, 
so that variation in model outputs is induced exclusively by the applied intervention operator $\mathcal{I}$.

For any input $x$ and intervention $\mathcal{I}$, we define the interventional response difference
\[
\delta_\theta(x, \mathcal{I}) = f_\theta(x^{\mathcal{I}}) - f_\theta(x),
\]
where subtraction is interpreted component-wise in $\mathcal{Y}$. This quantity captures the relative change in model output induced by a controlled structural perturbation, with the original input serving as a reference point for contrastive evaluation.

\subsection{Structured Inputs and Matched Interventions}

The auditing framework assumes that inputs admit a structured representation that enables selective modification of identifiable components. Formally, each input $x \in \mathcal{X}$ is represented as a collection of components
\[
x = \big(x^{(1)}, x^{(2)}, \dots, x^{(M)}\big),
\]
where each component corresponds to a well-defined subset of the input representation. The nature of this decomposition is domain dependent and may correspond to residues in a biological sequence, nodes in a graph, regions in an image, or fields in a structured record. This decomposition is not assumed to reflect true causal structure.

An intervention acts on a subset of components indexed by a scope $\mathcal{S} \subseteq \{1, \dots, M\}$. For an input $x$ and intervention $\mathcal{I}$, the intervened input satisfies
\[
x^{\mathcal{I}}_{(m)} =
\begin{cases}
\phi(x^{(m)}) & \text{if } m \in \mathcal{S}, \\
x^{(m)} & \text{otherwise},
\end{cases}
\]
where $\phi$ denotes the intervention operator applied to the selected components. The operator specifies how components are modified, while the scope specifies where the modification occurs.

This separation induces matched intervention pairs with identical scope size and intervention operator, differing only in structural alignment.
For auditing to be well-defined, each input is required to admit a unique interventional representation, so that interventional effects are unambiguously associated with individual inputs.

\subsection{Mechanistic and Spurious Intervention Classes}

Reasoning behavior is evaluated by contrasting model responses to two classes of matched input-level interventions.
We assume access to an externally specified structural prior that identifies a subset of input components relevant to the prediction task.
The prior is used solely to define structural alignment and does not encode ground-truth causal structure.

Let $\mathcal{S}_{\mathrm{prior}} \subseteq \{1,\dots,M\}$ denote the set of input components
identified by the structural prior.
Mechanistic intervention scopes are defined as subsets
$S \subseteq \mathcal{S}_{\mathrm{prior}}$ with cardinality defined relative to
$|\mathcal{S}_{\mathrm{prior}}|$.

We define a complementary set of input components $\mathcal{S}_{\mathrm{comp}} \subseteq \{1,\dots,M\}\setminus\mathcal{S}_{\mathrm{prior}}$, disjoint from the prior-defined region.
For each mechanistic scope $S \subseteq \mathcal{S}_{\mathrm{prior}}$, a spurious scope $S' \subseteq \mathcal{S}_{\mathrm{comp}}$ is constructed by uniformly sampling components outside $\mathcal{S}_{\mathrm{prior}}$ with matched cardinality $|S'|=|S|$; by construction, such scopes are not supported by the structural prior.

For a fixed intervention operator $\phi$, the corresponding intervention classes are
\[
\mathcal{I}^{\mathrm{mech}} = \{(S, \phi) \mid S \subseteq \mathcal{S}_{\mathrm{prior}}\},
\]
\[
\mathcal{I}^{\mathrm{spur}} = \{(S', \phi) \mid S' \subseteq \mathcal{S}_{\mathrm{comp}}\}.
\]

This construction enables a controlled comparison of model sensitivity to structurally aligned versus misaligned perturbations, while holding the intervention operator and scope cardinality fixed.

\subsection{Quantifying Reasoning Consistency}
All reasoning metrics are defined with respect to the auditing set $\mathcal{D}$ and characterize model behavior under controlled interventional perturbations.

ISAAC evaluates reasoning by comparing model responses to matched mechanistic and spurious interventions applied to the same input. This contrast isolates sensitivity to structurally aligned components while holding the intervention operator and perturbation magnitude fixed.

For a reference input $x \in \mathcal{D}$, mechanistic and spurious interventions induce two collections of interventional response differences,
\[
\Delta^{\mathrm{mech}}(x) =
\{\delta_\theta(x, \mathcal{I}) \mid \mathcal{I} \in \mathcal{I}^{\mathrm{mech}}\},
\]
\[
\Delta^{\mathrm{spur}}(x) =
\{\delta_\theta(x, \mathcal{I}) \mid \mathcal{I} \in \mathcal{I}^{\mathrm{spur}}\}.
\]
These sets summarize how model outputs vary when structurally aligned or misaligned components of the input are selectively perturbed.

\paragraph{Reasoning Score (RS).}
We introduce the \emph{Reasoning Score (RS)} as a per-input measure of relative interventional preference. For a given input $x$, we summarize the typical response to each intervention class as
\[
m_{\mathrm{mech}}(x) = \mathrm{median}\!\left(\Delta^{\mathrm{mech}}(x)\right),
\]
\[
m_{\mathrm{spur}}(x) = \mathrm{median}\!\left(\Delta^{\mathrm{spur}}(x)\right),
\]
where the median summarizes the central tendency of signed interventional responses within each class.

The RS is defined as
\[
\mathrm{RS}(x) =
\frac{|m_{\mathrm{mech}}(x)|}
{|m_{\mathrm{mech}}(x)| + |m_{\mathrm{spur}}(x)|}
\;\in\; [0,1].
\]
If $|m_{\mathrm{mech}}(x)| = |m_{\mathrm{spur}}(x)| = 0$, we set $\mathrm{RS}(x)=0.5$.

RS quantifies the relative magnitude of the dominant central interventional effect
between mechanistic and spurious perturbations. As it is based on the median of signed interventional responses, large but heterogeneous responses with
opposing signs lead to small median shifts and consequently low RS values, capturing
weak alignment rather than low sensitivity. Higher RS values indicate a stronger
relative preference for mechanistically aligned perturbations, whereas lower values
indicate a stronger preference for misaligned perturbations.
RS is therefore most informative when interpreted comparatively across models
evaluated under identical intervention designs and predictive regimes.

RS uses $|m_{\mathrm{mech}}(x)|$ rather than $\mathrm{median}(|\Delta^{\mathrm{mech}}(x)|)$
because the sign of the logit shift is not uniformly interpretable across
inputs; the signed median captures directional coherence within each class,
and its magnitude reflects the dominant central effect.
Directional analyses are reported in Appendix~\ref{app:directional}.

Model-level reasoning behavior is summarized by averaging RS over the auditing set,
\[
\mathrm{RS}_\theta = \mathbb{E}_{x \sim \mathcal{D}}[\mathrm{RS}(x)].
\]

\paragraph{Geometric diagnostic metrics.}
While RS captures relative intervention preference at the level of individual inputs, it does not characterize the global structure or dispersion of interventional responses across the auditing set. We therefore introduce two complementary diagnostic metrics that describe the geometry of mechanistic and spurious response distributions at the model level.

Let
\[
\Delta^{\mathrm{mech}}_\theta = \bigcup_{x \in \mathcal{D}} \Delta^{\mathrm{mech}}(x),
\qquad
\Delta^{\mathrm{spur}}_\theta = \bigcup_{x \in \mathcal{D}} \Delta^{\mathrm{spur}}(x),
\]
denote the collections of all mechanistic and spurious interventional responses, respectively.

The \emph{Separation Coefficient} is defined as
\[
C_{\mathrm{sep}}(\theta) =
\frac{
\left|
\mathrm{median}\!\left(\Delta^{\mathrm{mech}}_\theta\right)
-
\mathrm{median}\!\left(\Delta^{\mathrm{spur}}_\theta\right)
\right|
}{
\mathrm{IQR}\!\left(\Delta^{\mathrm{mech}}_\theta\right)
},
\]
and quantifies the separation between typical mechanistic and spurious responses, normalized by the intrinsic variability of mechanistic effects. If $\mathrm{IQR}(\Delta^{\mathrm{mech}}_\theta)=0$, we set $C_{\mathrm{sep}}(\theta)=0$.

The \emph{Overlap Rate} is defined as the fraction of spurious responses that fall within the interquartile range of mechanistic responses,
\[
\mathrm{Overlap}(\theta) =
\mathbb{P}\!\left(
s \in [Q_{0.25}^{\mathrm{mech}}, Q_{0.75}^{\mathrm{mech}}]
\;\middle|\;
s \in \Delta^{\mathrm{spur}}_\theta
\right).
\]

Together, these metrics complement RS by characterizing the distributional structure of mechanistic and spurious interventional responses beyond central tendency.

\section{Experiments}

\subsection{Experimental Questions and Evaluation Protocol}

The experimental evaluation assesses whether ISAAC distinguishes models whose predictions are supported by structurally aligned signals from models relying on spurious correlations, while holding predictive performance fixed. The focus is therefore on interventional response organization rather than predictive accuracy alone.

We address three questions: 
(i) whether models respond differently to mechanistic versus spurious interventions on identical inputs; 
(ii) whether such differences are consistently captured at the model level by ISAAC reasoning scores and geometric separation metrics; and 
(iii) whether ISAAC metrics reveal differences in reasoning behavior that are not detectable via standard accuracy-based evaluation.

All experiments follow a two-stage protocol that separates predictive validation from interventional auditing.

Models are trained and selected using standard training and validation procedures. Predictive performance is evaluated on the test split solely to verify that all models operate within a comparable predictive regime, and is not used for ranking or comparison.

Interventional auditing is performed subsequently on a structurally valid subset of the test data. Models are treated as frozen predictors and probed through structured input-level interventions, without retraining, model selection, or label-based evaluation.

Reported results therefore characterize the organization of model responses under matched mechanistic and spurious interventions, conditional on comparable predictive performance.

\subsection{Auditing Setup: Task, Data, and Models}

We apply ISAAC to DTI prediction, a scientific machine learning task in which valid predictions are expected to depend on localized protein--ligand interactions rather than global sequence statistics. Sequence-based DTI benchmarks are known to admit shortcut solutions despite strong predictive performance, making them suitable for auditing interventional reasoning beyond accuracy \citep{ilyas2019adversarial, geirhos2020shortcut}.

Interaction prediction is formulated as a binary classification task. Each input is a pair $x=(d,t)$, where $d$ is a drug encoded as a SMILES string and $t$ is a protein target encoded as an amino acid sequence, with label $y \in \{0,1\}$ indicating interaction presence \citep{chen2020tcpi, zeng2021moltrans, bai2022drugban, gaoming2025tapb}. Models implement a deterministic scoring function $f_\theta(d,t) \in \mathbb{R}$. In ISAAC, auditing is performed directly on this raw score, independently of any decision threshold or calibration.

Experiments are conducted on the Davis benchmark, a standard dataset for kinase inhibitor interactions \citep{ozturk2018deepdta, chen2020graphdta, zeng2021moltrans, gaoming2025tapb}. We follow the train/validation/test splits released with the TAPB benchmark and apply them consistently across all models, as defined in the corresponding public repository \citep{gaoming2025tapb}.\footnote{\url{https://github.com/GaomingL1n/TAPB}} The TAPB benchmark specifies a random partition of the data into training, validation, and test sets according to a 7:1:2 ratio, which we adopt without modification across all models.

These splits are used exclusively for model training and validation, and to verify that all architectures operate within a comparable predictive regime. Interventional auditing is not performed on the raw splits directly, but on a structurally valid auditing set derived post-hoc from the test split, as described below.

We audit DeepDTA \citep{ozturk2018deepdta}, DeepConvDTI \citep{ingoo2019deep}, and TAPB \citep{gaoming2025tapb}, spanning convolutional, deeper target-side, and target-aware attention-based architectures with increasing structural bias. Models are selected to contrast inductive assumptions relevant to ISAAC’s interventional analysis rather than predictive performance, and are audited post-hoc from frozen checkpoints trained using original implementations and recommended, comparable training protocols. No model-specific tuning is performed during auditing.

\subsection{Intervention Design and Structural Validity}

ISAAC evaluates reasoning behavior by contrasting model responses to matched mechanistic and spurious interventions, following the general paradigm of perturbation-based analysis for evaluating trained neural networks \citep{Ivanovs2021perturbation}. 
In the DTI setting, each input is a drug--target pair $x=(d,t)$, and interventions act exclusively on the protein sequence $t$, while the drug $d$ is held fixed.
This reflects the assumption that valid interaction predictions depend on localized target-side structure rather than global sequence statistics, consistent with known shortcut behavior in sequence-based DTI models \citep{ilyas2019adversarial, geirhos2020shortcut}.

Intervention scopes are defined using an externally specified structural prior that identifies protein regions expected to be mechanistically relevant for binding. For kinase targets in the Davis benchmark, this prior is derived from KLIFS (Kinase–Ligand Interaction Fingerprints and Structures) binding-pocket annotations \citep{kooistra2015klifs}, inducing a set of residue indices $\mathcal{S}_{\mathrm{prior}} \subseteq \{1,\dots,M\}$. The prior is used solely to define intervention locations and does not encode ground-truth causal effects.

Auditing is performed on a structurally valid auditing set derived post-hoc from the test split. Out of 379 protein targets in the Davis test split, 321 (84.7\%) have KLIFS binding-pocket annotations, and 208 (54.9\%) admit fully realizable interventions, meaning that all annotated pocket residues can be unambiguously indexed in the corresponding protein sequences. This deterministic filtering ensures that each audited input admits a unique and well-defined interventional representation. Structural coverage and key properties of the intervention design are summarized in \cref{tab:intervention_design}.

\begin{table}[t]
\centering
\caption{
Structural coverage and intervention design statistics for the Davis benchmark.
Reported values correspond to successive structural requirements needed to define unambiguous input-level interventions under ISAAC.
}
\label{tab:intervention_design}
\begin{tabular}{ll}
\toprule
\multicolumn{2}{l}{Structural requirements} \\
\midrule
Targets in test split & 379 \\
Targets with KLIFS pocket annotation & 321 (84.7\%) \\
Targets with realizable interventions & 208 (54.9\%) \\
\midrule
\multicolumn{2}{l}{Mechanistic region size (KLIFS)} \\
\midrule
Median pocket residues per target & 85 \\
IQR pocket residues per target & 8 \\
\midrule
\multicolumn{2}{l}{Intervention control} \\
\midrule
Exact cardinality matching (mech./spur.) & 100\% \\
\bottomrule
\end{tabular}
\end{table}

For each input, mechanistic scopes are generated by sampling fixed-size subsets $\mathcal{S} \subseteq \mathcal{S}_{\mathrm{prior}}$ from the prior-defined binding-pocket region. Intervention cardinality is defined relative to target-specific pocket size.

For each sampled mechanistic scope, a matched spurious scope $\mathcal{S}'$ of
identical cardinality is constructed by uniformly sampling residues from
$\{1,\dots,M\}\setminus\mathcal{S}_{\mathrm{prior}}$. This enforces exact
matching of perturbation magnitude, so that the two intervention classes differ
solely in their alignment with the structural prior. By construction, however,
mechanistic and spurious scopes also differ in geometric structure: mechanistic
scopes are spatially compact (mean positional spread 227, contiguity 0.922),
whereas spurious scopes are more diffuse (mean spread 754, contiguity 0.195).
The implications of this asymmetry are assessed in
Appendix~\ref{app:locality}.

We consider a family of input-level intervention operators that selectively perturb local structural information while preserving global sequence properties, instantiated in our experiments via masking and physicochemically constrained substitutions.
All interventions are defined independently of the model and applied identically across architectures, ensuring that audited differences reflect model behavior under a shared intervention design rather than dependence on a specific perturbation choice.

\subsection{Auditing Results}

We analyze how trained DTI models respond to matched mechanistic and spurious interventions under the ISAAC auditing protocol. The goal is to characterize differences in interventional reasoning behavior across models, conditional on operating within a comparable predictive regime.

\paragraph{Predictive performance comparison.}
Predictive accuracy is used as a minimal validation criterion to ensure that
audited models operate within a comparable and well-defined predictive regime;
performance metrics serve as a reality check rather than an explanatory
measure of model behavior \citep{YuKumbier2020VeridicalDataScience}.
\cref{tab:auroc_control} reports AUROC on the auditing subset (208 targets,
3{,}044 drug--target pairs), averaged across 5 training seeds, to verify that
the comparable-regime assumption holds specifically on the data used for
interventional analysis.

\begin{table}[t]
\centering
\caption{
Predictive performance of the audited models on the auditing subset
(208 targets with fully realizable ISAAC interventions). Values are means with 95\% confidence intervals across 5 training seeds.
}
\label{tab:auroc_control}
\begin{tabular}{lc}
\toprule
Model & AUROC (95\% CI) \\
\midrule
DeepConvDTI & 0.876 (0.874, 0.879) \\
DeepDTA     & 0.907 (0.900, 0.917) \\
TAPB        & 0.882 (0.833, 0.909) \\
\bottomrule
\end{tabular}
\end{table}

All models achieve nearly indistinguishable predictive performance on the auditing subset, with absolute AUROC differences around 3\%, confirming that
observed interventional differences are not attributable to residual
performance gaps on the analyzed data.

\paragraph{Aggregated reasoning metrics.}
We quantify model-level reasoning behavior using the ISAAC metrics averaged
across 5 independent runs, varying both training and intervention seeds.
Reporting multi-seed averages ensures that observed differences reflect stable
properties of each architecture rather than artifacts of a single training or
sampling realization. These metrics characterize interventional response
structure and are not intended as alternative performance measures.
\cref{tab:auditing_results} summarizes the resulting aggregated scores.

\begin{table*}[t]
  \centering
  \caption{
  Auditing results under structured interventions on the Davis benchmark,
  averaged across 5 runs varying both training and intervention seeds.
  RS denotes the Reasoning Score averaged over the auditing set.
  $C_{\mathrm{sep}}$ quantifies geometric separation between mechanistic and
  spurious interventional responses, and Overlap measures their distributional
  overlap. Values are reported as means with 95\% confidence intervals obtained
  from a hierarchical bootstrap resampling inputs and interventional responses,
  averaged across runs.
  }
  \label{tab:auditing_results}
  \begin{tabular}{lccc}
    \toprule
    Model & RS (95\% CI) & $C_{\mathrm{sep}}$ (95\% CI) & Overlap (95\% CI) \\
    \midrule
    DeepConvDTI & 0.414 (0.406, 0.421) & 0.061 (0.028, 0.100) & 0.347 (0.329, 0.366) \\
    DeepDTA     & 0.441 (0.432, 0.450) & 0.079 (0.037, 0.122) & 0.379 (0.358, 0.402) \\
    TAPB        & 0.521 (0.512, 0.530) & 0.458 (0.417, 0.497) & 0.417 (0.394, 0.444) \\
    \bottomrule
  \end{tabular}
\end{table*}

Despite nearly identical AUROC, the audited models exhibit consistent and
reproducible differences in the organization of their interventional responses
across all runs. TAPB achieves the highest mean RS (0.521, 95\%~CI
[0.512,~0.530]) together with a substantially larger separation coefficient
($C_{\mathrm{sep}} = 0.458$, 95\%~CI [0.417,~0.497]), indicating coherent
separation between mechanistic and spurious interventional responses under the
auditing protocol. In contrast, DeepConvDTI exhibits the lowest RS
(0.414, 95\%~CI [0.406,~0.421]) and a near-zero separation coefficient
($C_{\mathrm{sep}} = 0.061$, 95\%~CI [0.028,~0.100]), reflecting weak and
diffuse sensitivity with no coherent preference for mechanistically aligned
perturbations. DeepDTA attains an intermediate RS (0.441, 95\%~CI
[0.432,~0.450]) with similarly low separation, indicating increased
responsiveness without coherent structural organization.

Notably, among models whose AUROC differs by approximately 3\%, model-level RS
values span a range from 0.414 to 0.521. This discrepancy
is stable across seeds and intervention families (see Appendix~\ref{app:robustness}), indicating
that accuracy-based evaluation alone does not reflect differences in the
structural organization of interventional responses captured by ISAAC.

Taken together, these results show that predictive performance alone is
insufficient to characterize how DTI models organize their responses under
intervention. ISAAC reveals systematic and reproducible differences in the
relative alignment and separation of interventional responses across models
operating within the same predictive regime, differences that remain invisible
under standard accuracy-based evaluation.

\paragraph{Reproducibility.}
Code will be made available upon publication.

\section{Discussion}

The central finding of this work is that predictive accuracy is insufficient
to characterize how trained DTI models utilize structured input information
under intervention. Models with AUROC differences around 3\% on the auditing subset, exhibit substantial and reproducible differences in RS, stable across 5 independent runs varying both training and
intervention seeds. This stability indicates that the observed differences
reflect genuine architectural properties rather than artifacts of a single
training realization. The dissociation is further reinforced by the AUROC
ordering being discordant with the ISAAC ordering: the model with the highest
audited-subset AUROC does not exhibit the strongest interventional alignment,
confirming that residual performance differences cannot account for the
observed behavior.

These differences are not arbitrary. TAPB, the only architecture with explicit target-aware attention, is the only model with mean values above the neutral threshold on all directional metrics (see 
Appendix~\ref{app:directional}): mechanistic dominance
above 0.5, mechanistic-to-spurious ratio above 1, and sign consistency of
0.741. DeepDTA and DeepConvDTI, which rely on convolutional encoding without
explicit binding-site awareness, remain below these thresholds, responding
more strongly to diffuse spurious perturbations than to compact mechanistic
ones. This pattern suggests that architectural inductive biases toward
localized target-side structure translate into measurably different
interventional behavior, even when predictive performance is indistinguishable.

This contribution is complementary to, and distinct from, existing evaluation
approaches. Feature attribution methods provide correlational explanations of
individual predictions but cannot distinguish mechanistic from spurious
sensitivity. Invariance-based approaches such as IRM address training
objectives rather than auditing the behavior of frozen models. Causal
abstraction and mediation analysis frameworks probe internal model components
but require access to model internals and meaningful decompositions of latent
representations. ISAAC instead operates post-hoc at the input level, requires
no retraining or internal access, and provides a controlled interventional
contrast under a shared structural prior, a combination not available in
existing frameworks.

The present study has several limitations that define its scope and motivate
future extensions. The structural prior is derived from KLIFS pocket
annotations and may not capture all affinity-determining residues; metrics
should therefore be interpreted as reflecting prior-relative structural
sensitivity rather than mechanistic validity in a strong causal sense. The
spurious control is matched in cardinality but not in spatial contiguity,
introducing a geometric asymmetry whose implications are assessed in
Appendix~\ref{app:locality}; geometry-matched controls represent an important
extension of the current design. The audit is restricted to target-side
interventions with the drug held fixed, and to three sequence-based
architectures on a single dataset. Extensions to drug-side interventions,
graph-based or structure-based model classes, and additional datasets such
as KIBA are compatible with the ISAAC framework given an appropriate
structural prior, and constitute the most direct directions for future work.

Overall, these results establish that post-hoc auditing of prior-relative
structural sensitivity provides information about model behavior that
accuracy-based evaluation cannot detect. ISAAC introduces a principled and
reproducible protocol for this analysis, applicable across architectures and
domains where structured inputs and externally specified priors are available.
Where predictive benchmarks are saturated, interventional auditing of this
kind offers a principled basis for distinguishing models that are
indistinguishable under conventional evaluation.
\section*{Impact Statement}

This work contributes to the evaluation of deep learning systems by proposing a structured approach for auditing interventional reasoning behavior beyond predictive accuracy.
ISAAC provides a principled way to analyze how trained models respond to controlled input perturbations, even in settings where standard performance metrics are saturated and fail to distinguish model behavior.

For the machine learning community, ISAAC highlights the distinction between predictive success and the organization of reasoning under intervention.
By conditioning on comparable accuracy and contrasting responses to structurally aligned and misaligned perturbations, the framework enables systematic comparison of models that are indistinguishable under conventional benchmarks.

More broadly, this work contributes to the practice of post-hoc model auditing by operationalizing interventional analysis without requiring retraining, intervention-level supervision, or access to internal model states.
ISAAC is model-agnostic and compatible with existing training pipelines, making it applicable across architectures and application domains where structured inputs are available.

By encouraging evaluation protocols that probe model behavior under meaningful perturbations, this work supports more transparent and informative assessment practices in scientific machine learning and related areas where understanding model behavior is critical alongside predictive performance.


\newpage
\appendix
\onecolumn
\section{Robustness Analyses}
\label{app:robustness}

This appendix provides additional analyses of the \textsc{ISAAC} audit on the
Davis benchmark. The analyses address three aspects of the empirical protocol:
the dependence of the Reasoning Score on the intervention operator, the
directional organization of interventional responses, and the geometric
properties of the mechanistic and spurious intervention scopes.

\paragraph{Operator-specific robustness.}\label{app:operators}

We evaluate the Reasoning Score (RS) under two intervention operators: masking
and physicochemically constrained substitution. Masking replaces selected
residues with a dedicated mask token (\texttt{X}), removing local residue
identity while preserving sequence length. Physicochemically constrained
substitution replaces each selected residue with an amino acid sampled from the
same physicochemical class, preserving coarse biochemical properties while
perturbing residue identity. For both operators, intervention realizations are
fixed across architectures by using the same random seeds for scope sampling and
substitution.

RS is computed separately for each operator and averaged across 5 independent
runs varying both model-training and intervention seeds. The results are shown in
\cref{tab:robustness_operators}. Absolute RS values differ between operators,
as expected given their different perturbation semantics. The ordering of the
models, however, is unchanged under both operators. Thus, the qualitative
comparison between architectures is stable with respect to this operator choice.

\begin{table}[ht]
  \centering
  \caption{
  Operator-specific \textsc{ISAAC} Reasoning Scores on the Davis benchmark,
  averaged across 5 independent training and intervention seeds.
  $\text{RS}_{\mathrm{mask}}$ and $\text{RS}_{\mathrm{sub}}$ denote RS under
  masking and physicochemically constrained substitution, respectively. Values
  are means with 95\% confidence intervals obtained by hierarchical bootstrap,
  averaged across runs.
  }
  \label{tab:robustness_operators}
  \begin{tabular}{lcc}
    \toprule
    Model
      & $\text{RS}_{\mathrm{mask}}$ (95\% CI)
      & $\text{RS}_{\mathrm{sub}}$ (95\% CI) \\
    \midrule
    DeepConvDTI & 0.402 (0.394, 0.410) & 0.427 (0.419, 0.433) \\
    DeepDTA     & 0.427 (0.419, 0.436) & 0.448 (0.439, 0.456) \\
    TAPB        & 0.440 (0.430, 0.449) & 0.586 (0.577, 0.595) \\
    \bottomrule
  \end{tabular}
\end{table}

\paragraph{Directional organization of interventional responses.}\label{app:directional}

The RS summarizes the relative magnitude of central responses to mechanistic and
spurious perturbations. It does not directly characterize whether responses
within an intervention class are directionally coherent. We therefore consider
three complementary diagnostics.

\emph{Sign Consistency} (SC) is the fraction of mechanistic responses whose sign
matches the sign of the within-input central mechanistic response
$m_{\mathrm{mech}}(x)$, averaged over the auditing set. Values above $0.5$
correspond to directional agreement within mechanistic perturbations.

\emph{Mechanistic-to-Spurious Ratio} (MSR) is the average of
$|m_{\mathrm{mech}}(x)|/|m_{\mathrm{spur}}(x)|$, with the convention
$\mathrm{MSR}=1$ when both central responses are zero. Values above $1$ indicate
larger central responses to mechanistic than to matched spurious perturbations.

\emph{Mechanistic Dominance} (MD) is the fraction of audited inputs satisfying
$|m_{\mathrm{mech}}(x)| > |m_{\mathrm{spur}}(x)|$. Values above $0.5$ indicate
that mechanistic perturbations produce the larger central response for a
majority of inputs.

The results are reported in \cref{tab:directional}. TAPB has the largest SC and
is the only model with mean MSR above $1$ and mean MD above $0.5$. DeepDTA and
DeepConvDTI remain below these thresholds for MSR and MD, indicating larger
central responses to spurious perturbations for most audited inputs. TAPB's MD
has higher across-run variability ($0.535 \pm 0.114$), so the dominance result is
best interpreted as an aggregate pattern rather than a uniform per-run effect.

\begin{table}[ht]
  \centering
  \caption{
  Directional organization of interventional responses on the Davis benchmark,
  averaged across 5 runs varying both training and intervention seeds.
  SC: Sign Consistency; MSR: Mechanistic-to-Spurious Ratio; MD: Mechanistic
  Dominance, defined as the fraction of inputs satisfying
  $|m_{\mathrm{mech}}| > |m_{\mathrm{spur}}|$. Values are reported as
  mean $\pm$ standard deviation across runs.
  }
  \label{tab:directional}
  \begin{tabular}{lccc}
    \toprule
    Model & SC (mean $\pm$ std) & MSR (mean $\pm$ std) & MD (mean $\pm$ std) \\
    \midrule
    DeepConvDTI & $0.524 \pm 0.012$ & $0.665 \pm 0.069$ & $0.328 \pm 0.046$ \\
    DeepDTA     & $0.554 \pm 0.015$ & $0.754 \pm 0.058$ & $0.398 \pm 0.027$ \\
    TAPB        & $0.741 \pm 0.050$ & $1.245 \pm 0.323$ & $0.535 \pm 0.114$ \\
    \bottomrule
  \end{tabular}
\end{table}

\paragraph{Geometry of intervention scopes.}\label{app:locality}

Mechanistic and spurious scopes are exactly matched in cardinality but not in
spatial geometry. Mechanistic scopes are sampled from the KLIFS-defined
binding-pocket region, whereas spurious scopes are sampled outside that region.
Consequently, mechanistic scopes are more compact, while spurious scopes are
more diffuse. \Cref{tab:locality} reports the corresponding descriptive
statistics.

\begin{table}[ht]
  \centering
  \caption{
  Geometric properties of mechanistic and spurious intervention scopes on the
  Davis auditing set. Positional spread is the mean sequence-position distance
  between the first and last selected residue. Contiguity is the fraction of
  selected residues adjacent to at least one other selected residue, averaged
  over scopes.
  }
  \label{tab:locality}
  \begin{tabular}{lcc}
    \toprule
    Scope type & Mean positional spread & Mean contiguity \\
    \midrule
    Mechanistic & 227 & 0.922 \\
    Spurious    & 754 & 0.195 \\
    \bottomrule
  \end{tabular}
\end{table}

This geometric asymmetry defines the scope of the present control. The
mechanistic--spurious comparison is cardinality-matched and operator-matched,
but not geometry-matched. The audit should therefore be interpreted as a contrast
between prior-aligned and prior-misaligned perturbations under matched
perturbation size, rather than as a test in which all geometric properties are
held fixed.

Within this design, the response patterns differ across architectures.
DeepConvDTI and DeepDTA show larger central responses to the more diffuse
spurious scopes, with MSR below $1$ and MD below $0.5$. TAPB shows the opposite
aggregate pattern, with MSR above $1$ and MD above $0.5$, although MD remains
variable across runs. These results do not rule out a contribution of locality,
but they show that the geometric asymmetry does not impose a uniform response
pattern across models. A fully geometry-matched control is therefore a natural
extension of the present audit, while the current results remain consistent with
architecture-dependent sensitivity to the prior-defined binding-pocket region.

\end{document}